\pdfoutput=1
\documentclass[10pt,twocolumn,letterpaper]{article}

\usepackage{cvpr}
\usepackage{times}
\usepackage{epsfig}
\usepackage{graphicx}
\usepackage{amsmath}
\usepackage{amssymb}
\usepackage{enumitem}
\usepackage{multirow}
\usepackage{caption} 
\usepackage[pagebackref=true,breaklinks=true,letterpaper=true,colorlinks,bookmarks=false]{hyperref}

\cvprfinalcopy 

\def\cvprPaperID{****} 

\ifcvprfinal\fi
\begin{document}

\title{Pathologist-level Classification of Histologic Patterns on Resected Lung Adenocarcinoma Slides with Deep Neural Networks}

\author{\normalsize Jason W. Wei,$^{1,2}$\thanks{jason.20@dartmouth.edu}  \hspace{0.1cm} Laura J. Tafe,$^3$ Yevgeniy A. Linnik,$^3$ Louis J. Vaickus,$^3$ Naofumi Tomita,$^1$ Saeed Hassanpour$^{1,2}$\thanks{saeed.hassanpour@dartmouth.edu} \\[1\baselineskip]
\normalsize $^1$Department of Biomedical Data Science, Dartmouth College\\
\normalsize $^2$Department of Computer Science, Dartmouth College\\
\normalsize $^3$Department of Pathology and Laboratory Medicine, Dartmouth-Hitchcock Medical Center\\
\normalsize $^4$Department of Epidemiology, Dartmouth College
}

\maketitle

\begin{abstract}
Classification of histologic patterns in lung adenocarcinoma is critical for determining tumor grade and treatment for patients. However, this task is often challenging due to the heterogeneous nature of lung adenocarcinoma and the subjective criteria for evaluation. In this study, we propose a deep learning model that automatically classifies the histologic patterns of lung adenocarcinoma on surgical resection slides. Our model uses a convolutional neural network to identify regions of neoplastic cells, then aggregates those classifications to infer predominant and minor histologic patterns for any given whole-slide image. We evaluated our model on an independent set of 143 whole-slide images. It achieved a kappa score of 0.525 and an agreement of 66.6\% with three pathologists for classifying the predominant patterns, slightly higher than the inter-pathologist kappa score of 0.485 and agreement of 62.7\% on this test set. All evaluation metrics for our model and the three pathologists were within 95\% confidence intervals of agreement. If confirmed in clinical practice, our model can assist pathologists in improving classification of lung adenocarcinoma patterns by automatically pre-screening and highlighting cancerous regions prior to review. Our approach can be generalized to any whole-slide image classification task, and code is made publicly available at {\small \url{https://github.com/BMIRDS/deepslide}}.
\end{abstract}

\section{Introduction}
Lung carcinoma is the leading cause of cancer death among both men and women in the United States and the western world.$^1$ It is classified into small cell neuroendocrine carcinoma and non-small cell carcinoma, of which adenocarcinoma is the most common histologic type, accounting for about half of all cases.$^2$ Treatment for lung adenocarcinoma is based on the grade and stage of the tumor, which is determined largely by pathologists' evaluation of the tumor's histology. In 2015, the World Health Organization released its most recent guidelines for the classification of non-mucinous lung adenocarcinoma, identifying five histologic patterns (subtypes): lepidic, acinar, papillary, micropapillary, and solid.$^{3,4}$ Furthermore, these guidelines recommend comprehensive documentation of minor components in addition to the predominant subtype, since lung adenocarcinomas frequently contain a heterogenous mix of multiple patterns.

Classification of adenocarcinoma histology patterns is important because it provides significant insight into patient prognosis and survival. For instance, identification of pure lepidic pattern has been shown to have excellent prognoses for stage I lung cancer patients and is typically classified as low grade.$^{5-9}$ Acinar and papillary patterns are classified as intermediate grade, while micropapillary and solid patterns are high grade and are associated with poor prognoses.$^{6,10-13}$ Patients with micropapillary or solid predominant patterns have lower survival rates, so they are more likely to undergo and benefit from adjuvant chemotherapy.$^{14}$ 

\begin{figure*}[htb]
\begin{centering}
\includegraphics[width=0.9\linewidth]{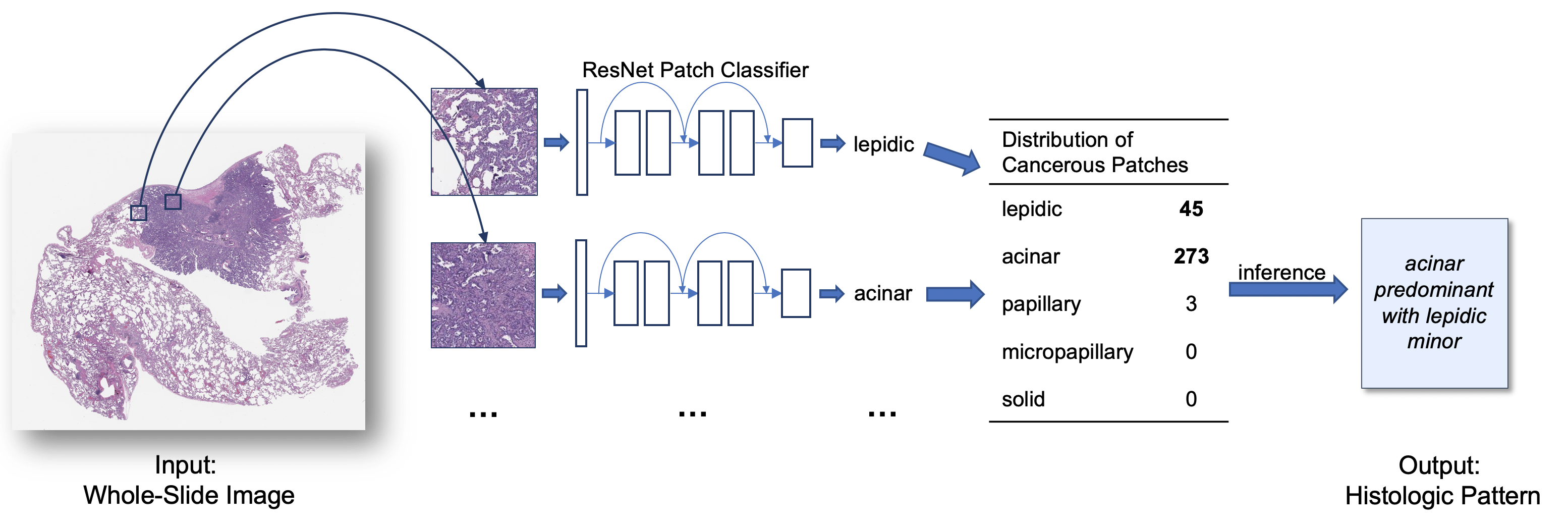}
\caption{Overview of whole-slide classification of histologic patterns. We used a sliding window approach on the whole slide to generate small patches, classified each patch with a residual neural network, aggregated the patch predictions, and used a heuristic to determine predominant and minor histologic patterns for the whole slide. Patch predictions were made independently of adjacent patches and relative location in the whole-slide image.} 
\label{fig:f1}
\end{centering}
\end{figure*}

Identifying the histologic subtypes in adenocarcinoma is extremely important for tumor prognosis and treatment, but accurate classification of such patterns can be challenging. About 80\% of adenocarcinoma cases contain a mixed spectrum of multiple histologic patterns,$^{15}$ and the qualitative criteria used for classification tends to induce inter-observer variability among pathologists. One study found that different appraisals of the maintenance or loss of alveolar structures resulted in varying classifications of acinar and lepidic patterns, and that major-minor classification of papillary and micropapillary subtypes was contentious when the two were intermixed.$^{16}$ Moreover, even small amounts of high grade patterns that are easily overlooked have been shown to be associated with significantly worse prognoses, especially for the micropapillary subtype.$^{17,18}$ The subjective nature of classifying such patterns has motivated several studies on the agreement of histologic subtype classifications among pathologists. One report revealed moderate to good kappa scores of 0.44-0.72 among pulmonary pathologists, and fair kappa scores of 0.38-0.47 among residents; intra-observer variability yielded kappa scores of 0.79-0.87.$^{19}$ Another survey of expert pulmonary pathologists found kappa scores of 0.70-0.84 for classical images and kappa scores of 0.24-0.52 for difficult cases.$^{20}$ 

Recent advances in artificial intelligence, particularly in the field of deep learning, have produced a set of image analysis techniques that automatically extract relevant features using a data-driven approach. One class of these deep learning models is convolutional neural networks, which have enabled researchers to create compelling algorithms for medical image analysis.$^{21,22}$ For pulmonary disease in particular, deep learning has already shown potential to assist pathologists in chest x-ray analysis, interstitial lung disease classification, and nodule detection.$^{23}$ The presented study is the first attempt to use emerging deep learning technology for automated classification of histologic subtypes on lung adenocarcinoma surgical resection slides.

\section{Results}
\noindent\textbf{A deep learning model for classification of whole-slide images.} This study presents a deep learning model for automated classification of histologic subtypes on lung adenocarcinoma histopathology slides. Our model uses a patch classifier combined with a sliding window approach to identify both major and minor patterns on a given whole-slide image, as shown in Figure \ref{fig:f1}. We used 422 whole-slide images collected from the Dartmouth-Hitchcock Medical Center in Lebanon, NH, which were randomly split into three sets: training, development, and test (Table \ref{tab:dist}). For final evaluation, we compared our model's classification of 143 whole-slide images in the independent test set to those of three pathologists.

\begin{table}[ht]
\scriptsize
\center
\begin{tabular}{l | c c c c c c}
\hline
 & \multicolumn{2}{c}{Training Set} & \multicolumn{2}{c}{Development Set} & Test Set & Total \\ \hline
Pattern & WSI & Crops & WSI & Patches & WSI & WSI \\ \hline\hline
Lepidic & 99 & 515 & 17 & 58 & 64 & 180 \\
Acinar & 115 & 692	& 23 & 269 & 82 & 220 \\
Papillary & 9 & 44 & 3 & 65 & 5 & 17 \\
Micropapillary & 41 & 412 & 9 & 50 & 22 & 72 \\
Solid & 68 & 425 & 9 & 400 & 54 & 131 \\
Benign & - & 2,073 & - & 226 & - & - \\ \hline
Total & 245 & 4,161 & 34 & 1,068 & 143 & 422 \\ \hline
\end{tabular}
\caption{Distribution of training, development, and test set data among five histologic patterns and benign cases. WSI denotes whole slide image. Crops are variable length and width and annotated by pathologists, while patches are square and of fixed size, obtained from sliding a window over crops. The class distribution for WSI's in our test set are the average of the labels from three pathologists.}
\label{tab:dist}
\end{table}

\begin{figure*}[htb]
\begin{centering}
\includegraphics[width=\linewidth]{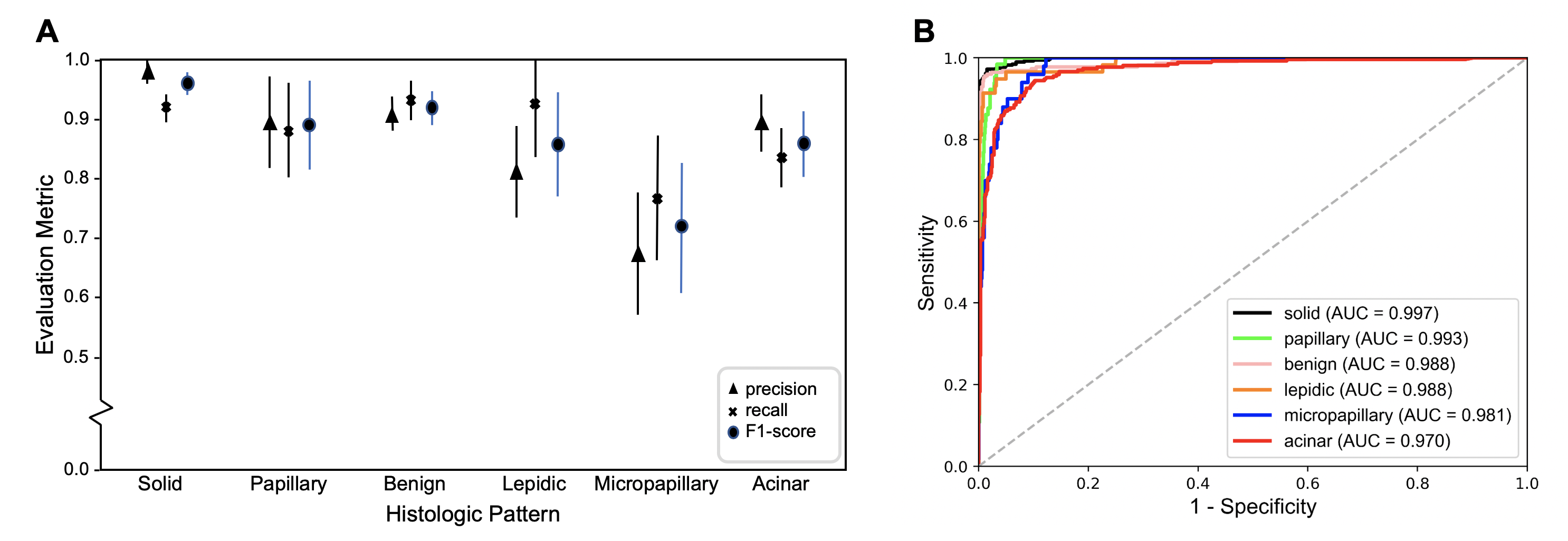}
\caption{Model’s performance on the 1,068 classic samples for histologic patterns. A: patch classification results with 95\% confidence intervals. B: ROC curves and their area under the curves (AUC’s) on this development set.} 
\label{fig:f2}
\end{centering}
\end{figure*}

\noindent\textbf{Accurate classification of classical examples of histologic subtypes.} For selection of the best neural network architecture, we validated our model against the development set of classic examples of histologic patterns. The best model achieved an F1 score of 90.4\% on this set of patches. The per-class evaluation metrics of precision, recall, and F1 score are shown in Figure \ref{fig:f2}A with their corresponding 95\% confidence intervals. In addition, we plotted the Receiver Operating Characteristic (ROC) curves of our model for each histologic class in Figure \ref{fig:f2}B. Our patch classifier achieved an area under the curve (AUC) greater than or equal to 0.97 for all classes.\\[1\baselineskip]
\begin{table}[ht]
\scriptsize
\center
\begin{tabular}{l | c c c}
\hline
& Kappa Score & Agreement (\%) & R. Agreement (\%) \\ \hline \hline
Pathologist 1 & 0.454 {\tiny (0.372-0.536)} & 61.3 {\tiny (53.3-69.3)} & 66.9 {\tiny (59.2-74.6)} \\
Pathologist 2 & 0.515 {\tiny (0.433-0.597)} & 64.8 {\tiny (57.0-72.6)} & 72.3 {\tiny (65.0-79.6)} \\
Pathologist 3 & 0.514 {\tiny (0.432-0.596)} & 63.1 {\tiny (55.2-71.0)} & 75.4 {\tiny (68.3-82.5)} \\
Inter-pathologist & 0.479 {\tiny (0.397-0.561)} & 62.7 {\tiny (54.8-70.6)} & 71.5 {\tiny (64.1-78.9)} \\
Baseline model$^{24}$ & 0.445 {\tiny (0.364-0.526)} & 60.1 {\tiny (52.1-68.1)} & 69.0 {\tiny (61.4-76.6)} \\
Our model & \textbf{0.525} {\tiny (0.443-0.607)} & \textbf{66.6} {\tiny (58.9-74.3)} & \textbf{76.7} {\tiny (69.8-83.6)} \\
 \hline
\end{tabular}\\
\caption{Comparison of pathologists and our model for classification of predominant subtypes in 143 whole-slide images in our test set. Average kappa score is calculated by averaging pairs of an annotator’s kappa scores. For instance, Pathologist 1 average is calculated by averaging the kappa scores of Pathologist 1 \& Pathologist 2, Pathologist 1 \& Pathologist 3, and Pathologist 1 \& our model. Average agreement was calculated in the same fashion. Robust agreement (R. Agreement)  indicates agreement for an annotator with at least two of the three other annotators. 95\% confidence intervals are shown in parentheses.}
\label{tab:results}
\end{table}

\begin{figure}[htb]
\begin{centering}
\includegraphics[width=\linewidth]{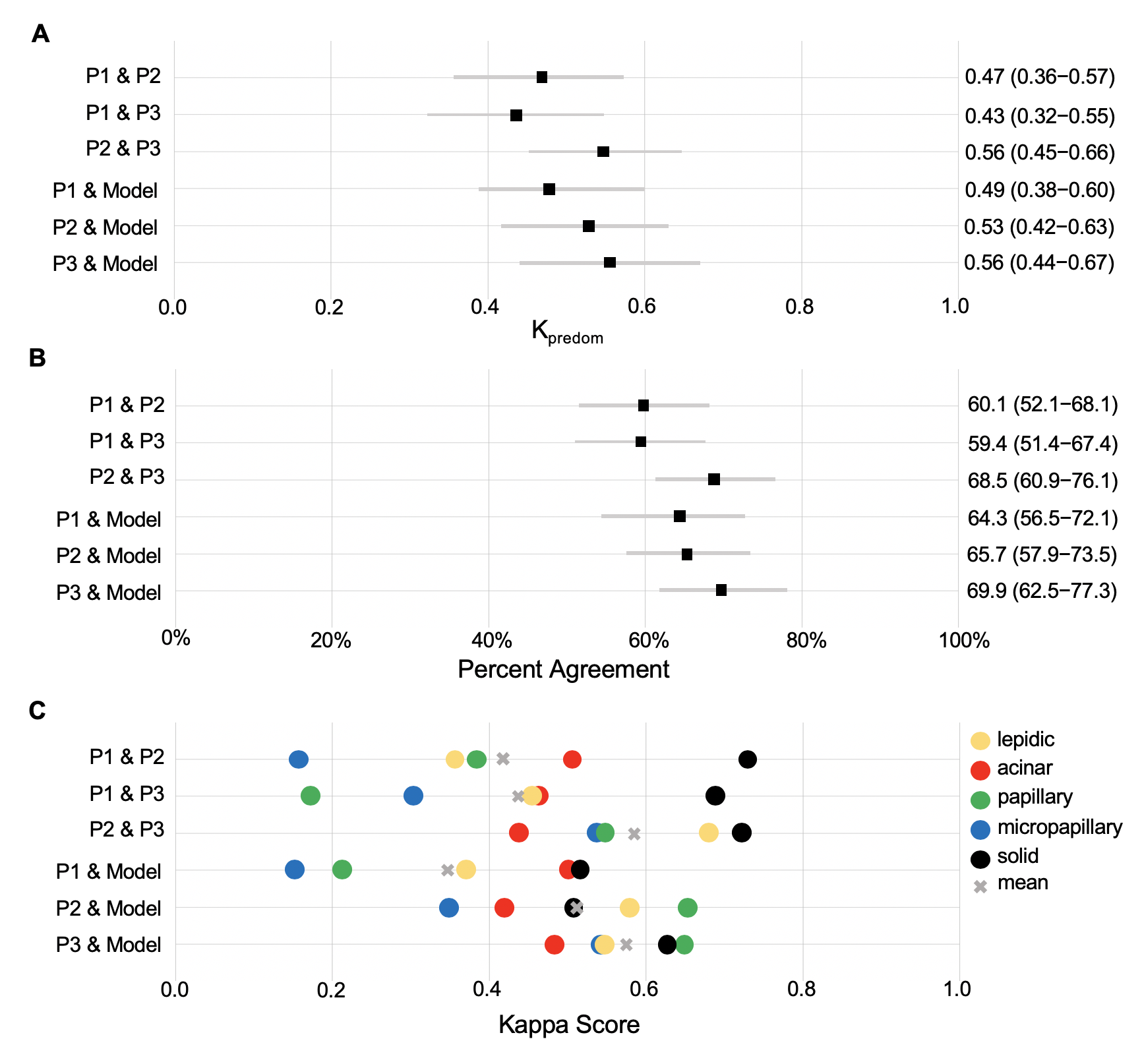}
\caption{Model's classification of 143 whole-slide images in the test set compared to those of three pathologists. A: the kappa score of the predominant classification among all pairs of annotations. B: agreement percentages of predominant classification among all pairs of annotations. C: kappa scores for each histologic pattern among all pairs of annotations regardless of predominant or minor subtypes. P1, P2, and P3 are Pathologist 1, Pathologist 2, and Pathologist 3 respectively. } 
\label{fig:f3}
\end{centering}
\end{figure}

\begin{figure*}[htb]
\begin{centering}
\includegraphics[width=0.78\linewidth]{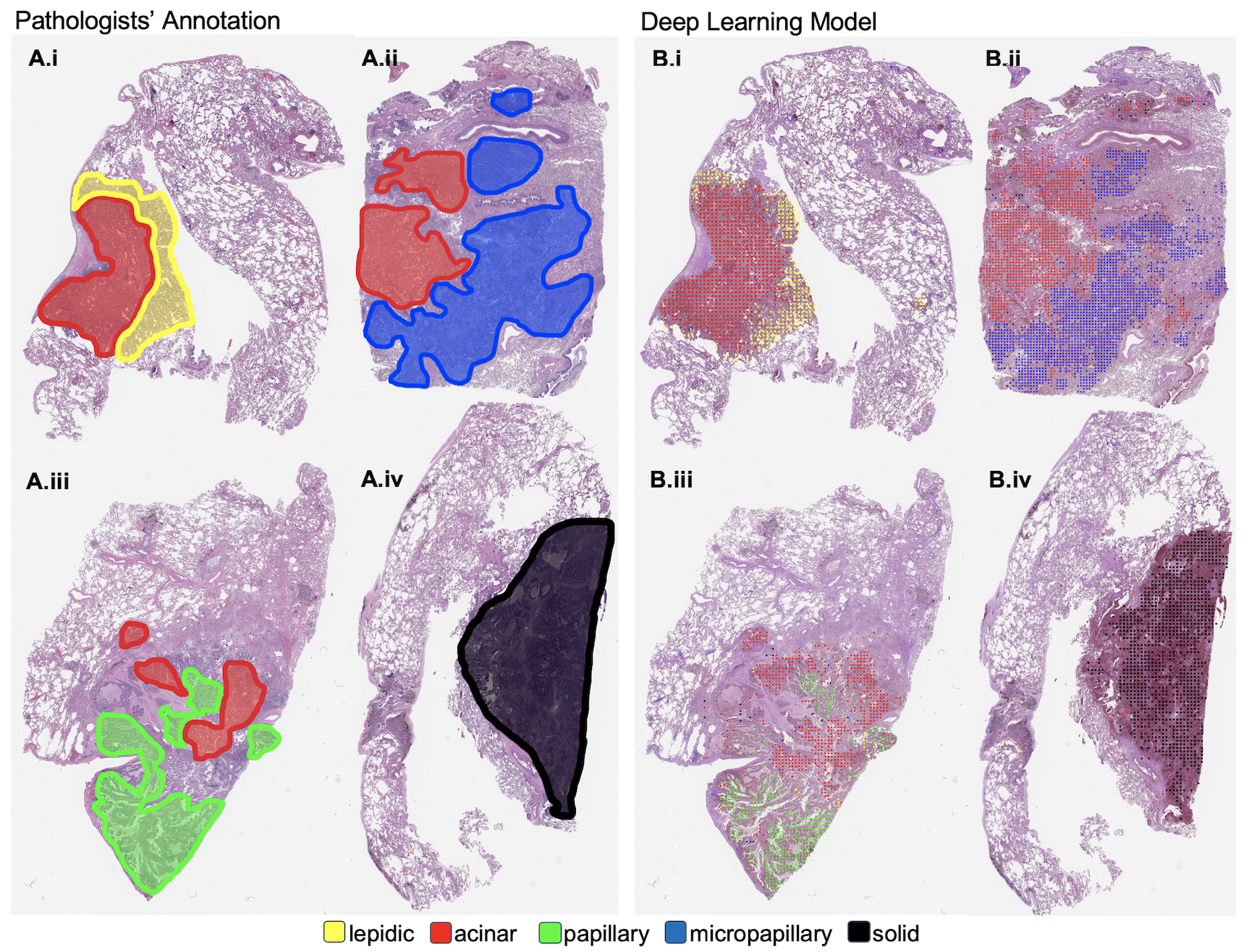}
\caption{Visualization of the histologic patterns annotated by pathologists (A.i-iv), compared to those detected by our model (B.i-iv).} 
\label{fig:f4}
\end{centering}
\end{figure*}

\noindent\textbf{Comparison of deep learning model to pathologists.} Our model was evaluated against pathologists on an independent test set of 143 whole-slide images. The kappa scores for predominant classification between every pair of annotators/model are shown in Figure \ref{fig:f3}A. In addition, Figure \ref{fig:f3}B shows the percentage of agreements on the predominant histologic patterns among our pathologist annotators and the final model. Figure \ref{fig:f3}C shows the kappa score for the detection of each histologic pattern, regardless of predominant or minor subtype. Table \ref{tab:results} summarizes the metrics in Figure \ref{fig:f3}A and Figure \ref{fig:f3}B through average kappa scores and average predominant agreement among the annotators/model. Notably, our model edges out inter-pathologist agreement measures with an average kappa score of 0.525 and an average predominant agreement of 66.6\%. Furthermore, for each annotator we calculated a metric called ``robust agreement", which indicates the annotator's agreement with at least two of the three other annotators. We performed a two-sample t-test for means on each pair of metrics in Figure \ref{fig:f3}A and Figure \ref{fig:f3}B, and found that our model and all pathologists' performances were within 95\% confidence intervals of agreement for every pair of metrics. For comparison, we also implemented the method used in Coudray et al. as a baseline.$^{24}$ Their method classified adenocarcinoma and squamous cell carcinoma slides by averaging the predicted probability of patches. We extended this methodology for a multi-label classification baseline in our study.  Finally, Figure \ref{fig:f4} depicts the visualization of the histologic patterns identified by our model on sample whole-slide images. A subjective qualitative investigation by our pathologist annotators confirmed that the patterns detected on each slide are on target.

\section{Discussion}
Our model is statistically on par with pathologists for all evaluated metrics on our test set of 143 whole-slide images. For all pairs of pathologists/model, $K_{predom}$ was in the moderate (0.41-0.60) range, with predominant agreement around sixty to sixty-five percent. Our model slightly edged out the pathologists on these two metrics, possibly because computing tumor areas by counting the number of patches is more precise than unaided estimations of tumor area by the naked eye. Of all disagreements in predominant subtype classification, 39.5\% were between the acinar and lepidic subtypes, a finding that is consistent with the fact that the two patterns often appear together, and it can be challenging to define an exact border between these two patterns. Detection of minor patterns, on the other hand, was more challenging both for pathologists and for our model. This is likely due to the fact that patterns that occur in small amounts can be interpreted differently or easily overlooked, leading to higher levels of disagreement among annotators. Our model was evaluated on an unbiased dataset from all available adenocarcinoma lobectomy slides since 2016 available at our institution and performs at least on par with pathologists in identifying the predominant and minor histologic subtypes. Of note, we are not aware of any other existing model for automated classification of lung adenocarcinoma patterns.

An automated system for detecting and visualizing histologic patterns of lung adenocarcinoma has a wide variety of applications in clinical settings. Considering the quick turnaround time of our model, it could be integrated into existing laboratory information management systems to automatically pre-populate diagnoses for histologic patterns on slides or provide a second opinion on more challenging patterns. In addition, a visualization of the entire slide, examined by our model at the piecewise level, could highlight elusive areas of high-grade patterns as well as primary regions of tumor cells. Also, our model could expedite the tumor diagnosis process by automatically requesting genetic testing for certain patients based on the histologic patterns detected, allowing clinicians to diagnose and treat patients faster. The application of our model in a clinical setting, which our research team will pursue as a next step, is essentially an automated platform for quality assurance in reading histologic slides of lung adenocarcinoma. A successful implementation of this system will support more accurate classification of lung cancer grade and ultimately facilitate the entire process of lung cancer diagnosis.

The model presented in this paper is rooted in strong deep learning methodology and achieves pathologist-level performance on the test set. However, one limitation is that our study is conducted on data from a single medical center, so our data may not necessarily be representative of all lung adenocarcinoma histology patterns. Although our whole-slide scans are of high resolution and we were able to use image augmentation to generate a large number of training samples, our dataset is relatively small in relation to classical deep learning datasets, many of which have more than ten thousand unique examples per class$^{25, 26}$  and more than a million unique images in total.$^{27}$ In particular, the papillary and micropapillary classes were extremely rare in our dataset, only represented in four and seventeen percent of the whole-slides images in our training set, respectively. Collection of more images through collaboration with other medical centers would allow us to refine our model on a more diverse dataset and will be pursued as future research.

Previous work has been done involving deep learning and lung cancer pathology images. In several studies, The Cancer Genome Atlas (TCGA) data was used to predict prognosis using computational methods.$^{28-30}$ While these papers have revealed meaningful correlations between tissue features and survival rates, their performances are not high enough to be used reliably in clinical practice. A recent study used TCGA data to predict mutations and distinguish between adenocarcinoma and squamous cell carcinoma.$^{24}$ Our work is novel in several ways. First, to the best of our knowledge, our study is the first to automate classification of histologic subtypes, a task that can be challenging even for experienced pathologists. Furthermore, we demonstrated a novel threshold-based aggregation method that yields performance surpassing those of previous studies,$^{24}$ allowing our model to be generalized to multi-class and multi-label tasks. Finally, while all previous work was done on frozen slides that are not typically used by pathologists for visual inspection, our model is trained on a comprehensively annotated set of formalin-fixed paraffin-embedded (FFPE) histopathology slides.

Moving forward, more work can be done to further the capabilities of our model. Possible improvements to our present architecture could include drawing bounding boxes around cancerous regions using region-based convolutional neural networks (R-CNN)$^{31}$ or outlining regions of interest at the pixel level using Mask-RCNN.$^{32}$ This would require a larger and more laboriously annotated dataset but could help pathologists recognize exactly which cellular structures the model identifies as cancerous. In addition, the predictions from our model can be tested for potential correlation with patient outcomes. Several previous studies have shown that even small amounts of the micropapillary pattern, which could be easily concealed on a whole-slide image, have been associated with extremely poor prognoses.$^{17,18}$ A study of our model's detection of micropapillary subtype compared to those of pathologists for patients who had unexpectedly worse survival rates could potentially shed insight on elusive histologic patterns easily missed by pathologists. In addition, studies have shown that certain histologic patterns are associated with specific mutations in EGFR, ALK, ROS1, and KRAS genes,$^{33-36}$ and that these mutations can be predicted by using deep neural networks on frozen slides.$^{37}$ With an appropriate dataset, our model could be re-trained to directly predict genetic mutations from FFPE slides and identify patients who require genetic screening and targeted therapy. To this end, we will continue our collaboration with the Pathology and Laboratory Medicine Department at our institution to retrieve the pathology reports and genetic screening results for the collected images in our current dataset, with the aim of training a model that can also predict genetic mutations and survival outcomes.

In this study, we proposed a deep learning model for classifying predominant and minor histologic patterns on lung adenocarcinoma whole-slide images. Our model consists of a residual convolutional neural network for patch classification combined with a whole-slide inferencing mechanism for determining predominant and minor subtypes on the whole slide. On an independent test set, our model performed on par with pathologists. The visualization of our results and a qualitative investigation by our pathologist annotators confirms that our model's classifications are generally on target. Our model can potentially be used to aid pathologists in classification of these histologic patterns and ultimately contribute to more accurate grading of lung adenocarcinoma. 

\section{Materials and Methods}
\noindent\textbf{Data Collection.} To develop and evaluate our model for classifying lung adenocarcinoma histology patterns, we collected whole-slide images from all patients with a diagnosis of lung adenocarcinoma since 2016 who underwent lobectomies at the Dartmouth-Hitchcock Medical Center (DHMC), a tertiary academic care center in Lebanon, New Hampshire. These histopathology slides contain formalin-fixed paraffin-embedded tissue specimens and were scanned by a Leica Aperio whole-slide scanner at 20x magnification at the Department of Pathology and Laboratory Medicine at DHMC. In total, 422 whole-slide images were collected for this study. We randomly partitioned 279 of these images (about two-thirds of the dataset) for model training, and the remaining 143 images (about one-third of the dataset) for model testing.\\[1\baselineskip]
\noindent\textbf{Slide Annotation.} All whole-slide images were manually labeled by three pathologists from the Department of Pathology and Laboratory Medicine at DHMC. The 279 images used for training were further split into a training set of 245 images and a development set of 34 images. For the training set, pathologists annotated 4,161 crops from 245 images, about 17 crops per image. These rectangular crops varied in size (mean: 718 $\times$ 771 pixels, standard deviation: 645 $\times$ 701 pixels, median: 429 $\times$ 473 pixels) and were labeled as either one of the five histologic patterns or benign. Our benign class also included inflammation, scarring, fibrosis, and artifacts. For the development set, our pathologists annotated 1,068 square patches of 224 $\times$ 224 pixels for classic examples of each pattern. Since these patches are used for model selection and development, all labels in this set were independently verified by two pathologists, and patches with disagreements were discarded. \\[1\baselineskip]
\noindent\textbf{Labeling the independent test set.} Our test set is 143 whole-slide images, each of which contains one or more of the five histological patterns. Our three pathologists independently labeled all images on the whole-slide level, specifying the predominant and minor patterns. After our model development was completed, we evaluated our model on this test set and compared its performance to those of our pathologist annotators. Table 1 shows the class distribution of crops for the training set, patches for the development set, and whole-slides for the test set.\\[1\baselineskip]
\noindent\textbf{Residual neural networks.} Deep learning models, such as convolutional neural networks, have been increasingly applied to computer vision and medical image analysis due to breakthroughs in high-performance computing and the availability of large datasets. In our study, we leverage the deep residual network (ResNet),$^{37}$ a type of convolutional neural network that uses residual blocks to achieve state-of-the-art performance on image recognition benchmarks such as ImageNet$^{38}$ and COCO.$^{39}$ We implemented ResNet to take in square patches as inputs and output a prediction probability for each of the five histological patterns and benign tissue, six classes in total. \\[1\baselineskip]
\noindent\textbf{Data processing and augmentation.} We trained our model on 4,161 annotated crops from the training set. Because each of these crops is of variable size, we used a sliding window algorithm to generate multiple smaller patches of fixed length and width from each crop. Some classes contained more crops than others, so we generated patches with different overlapping areas for each class to form a uniform class distribution. Before inputting a patch into the model for training, we normalized the color channel values to the mean and standard deviation of the entire training set to neutralize color differences between slides. Next, we augmented our training set by performing color jittering on the brightness, contrast, saturation, and hue of each image. Finally, we rotated each image by 90$^{\circ}$ and randomly flipped it across the horizontal and vertical axes. Our final training set consisted of approximately eight thousand patches per class.\\[1\baselineskip]
\noindent\textbf{Training the residual neural network.} As for model parameters, we initialized the network weights with the He initialization.$^{40}$ We trained for fifty epochs on the augmented training set, starting with an initial learning rate of 0.001 and decaying by a factor of 0.9 every epoch. Our model used the multi-class cross-entropy loss function. To find the optimal depth for the residual network, we conducted an ablation test on ResNets of 18, 34, 50, 101, and 152 layers. We found they all obtain similar accuracies on our development set, so we chose ResNet-18 since it has the smallest number of parameters and the fastest training time. Our final ResNet model for patch classification was trained in twenty-four hours on an NVIDIA K40c graphics processing unit (GPU) card. \\[1\baselineskip]
\noindent\textbf{Whole-slide inference.} At inference time, we aimed to detect all predominant and minor patterns at the whole-slide level. But because our trained ResNet classifies patches, not entire slides, we first broke down each whole slide into a collection of patches by sliding a fixed-size window over the entire image. Patches overlapped by a factor of one-fifth, resulting in a large number of patches for each high-resolution whole slide (mean = 9,267, standard deviation = 8,351, median = 7,069). We then classified each patch and filtered out noise by using thresholding to discard predictions of low confidence. Thresholds are determined by a grid search over each pattern class, optimizing for the correspondence between our model and whole-slide labels on the development set. Considering the distribution of the predicted patch patterns for each slide, we then used a three-step heuristic to classify the whole slide. First, classes comprising less than five percent of the patch predictions, as well as the benign class, were dropped. Then, the most frequent class was assigned to the predominant label. Finally, all remaining cancerous pattern classes were assigned to minor labels. By discarding predictions of low confidence and aggregating over a large number of patches, our model is robust to artifacts from tissue staining, as well as single-patch misclassifications. A schematic overview of the whole-slide inference process is shown in Figure 1. Evaluation time of our model for a single whole slide was around thirty seconds.\\[1\baselineskip]
\noindent\textbf{Statistical analysis and comparison to pathologists.} For final evaluation, we ran our model on the test set of 143 whole-slide images. We also asked our three pathologists to independently label the predominant and minor patterns in all 143 whole-slide images. As a result, we had four sets of whole-slide classifications in total: three from pathologists, and one from our model. To evaluate the performance of our model, we compared the concordance of our model's labels with those of pathologists' by calculating an interrater reliability metric called Cohen's kappa score.$^{41}$ We chose Cohen's kappa score for two reasons. First, because histologic patterns are only determined from subjective reviews by pathologists, no ground truth labels exist to calculate F1-scores or AUC. Second, previous studies on classifying histologic patterns use kappa score as a standard metric,$^{19, 20}$ so we follow this convention to facilitate comparison between our results and those of previous literature. Between every two sets of annotations, we calculated $K_{predom}$, predominant agreement, and kappa scores per class. $K_{predom}$ is the kappa score for the predominant pattern. Predominant agreement is the percentage of whole slides in which two annotators agreed on the predominant pattern. Kappa scores per class were calculated for detection of a pattern, regardless of predominant or minor subtype, between two sets of annotations. Furthermore, we calculated a metric called ``robust agreement", which indicates the agreement for an annotator with at least two of the three other annotators. We performed a two-sample t-test on all pairs of metrics described above to find any statistically significant difference among them.\\[1\baselineskip]
\noindent\textbf{Visualization of predicted patches.} We visualized the detected lung adenocarcinoma histologic patterns on whole-slide images by overlaying color-coded dots on patches for which our model predicted a histologic pattern. This visualization confirmed the decisions generated by our model and allowed pathologists to gain insight into our model's classification method. \\[1\baselineskip]
\noindent\textbf{Guidelines and regulations.} This study and the use of human subject data in this project were approved by the Dartmouth institutional review board (IRB) with a waiver of informed consent. The conducted research reported in this paper is in accordance with this approved IRB protocol and the World Medical Association Declaration of Helsinki on Ethical Principles for Medical Research Involving Human Subjects.

\section{Acknowledgements}
The authors would like to thank Laura Gordon and Arief Suriawinata, MD for their help with data collection for this study, and Lamar Moss and Sophie Montgomery for their feedback.

\section{Author Contributions}
All authors have read and approved the manuscript, and each author has participated sufficiently in developing the project and the manuscript. JW facilitated data collection and implemented the deep learning model. LT performed the literature search, advised study design, and contributed to data collection, data annotation, and evaluation of model's predictions. YL contributed to data collection, data annotation, and evaluation of model's predictions. LV contributed to data annotation and evaluation of the model's predictions. NT provided technical expertise for deep learning implementations. SH supervised the entire project and advised on study design, data collection, data annotation, and model implementation processes.

\section{Data Availability Statement}
The dataset used in this study is not publicly available due to patient privacy constraints. An anonymized version of this dataset can be generated and shared upon request from Saeed Hassanpour, PhD.

\section{References}
\begin{enumerate}[label={[\arabic*]}]
\item Torre LA, Siegel RL, Jemal A. Lung cancer statistics. Adv Exp Med Biol. 2015;893:1-19.
\item Meza R, Meernik C, Jeon J, Cote ML. Lung cancer incidence trends by gender, race, and histology in the United States. PLoS ONE. 2015;10:3.
\item Travis WD, Brambilla E, Nicholson AG, Yatabe Y, Austin JHM, Beasley MB, et al. The 2015 World Health Organization classification of lung tumors. J Thorac Oncol. 2015;9:1243-1260. 
\item Travis WD, et al. International association for the Study of Lung Cancer/American Thoracic Society/European Respiratory Society international multidisciplinary classification of lung adenocarcinoma. J Thorac Oncol. 2011;6:244-285.
\item Kadota K, et al. Prognostic significance of adenocarcinoma in situ, minimally invasive adenocarcinoma, and nonmucinous lepidic predominant invasive adenocarcinoma of the lung in patients with stage I disease. Am J Surg Pathol. 2014;38:448-460. 
\item Song Z, et al. Prognostic value of the IASLC/ATS/ERS classification in stage I lung adenocarcinoma patients-based on a hospital study in China. Eur J Surg Oncol. 2013;39:1262-1268.
\item Yoshizawa A, et al. Impact of proposed IASLC/ATS/ERS classification of lung adenocarcinoma: prognostic subgroups and implications for further revision of staging based on analysis of 514 stage I cases. Mod Pathol. 2011;24:653-664.
\item Warth A, et al. The novel histologic International Association for the Study of Lung Cancer/American Thoracic Society/European Respiratory Society classification system of lung adenocarcinoma is a stage-independent predict of survival. J Clin Oncol. 2012;30:1438-1446. 
\item Takahashi M, Shigematsu Y, Ohta M, Tokumasu H, Matsukura T, Hirai T. Tumor invasiveness as defined by the newly proposed IASLC/ATS/ERS classification has prognostic significant for pathologic stage IA lung adenocarcinoma and can be predicted by radiologic parameters. J Thorac Cardiovasc Surg. 2014;147:54-59.
\item Russell PA, Wainer Z, Wright GM, Daniels M, Conron M, Williams RA. Does lung adenocarcinoma subtype predict patient survival? A clinicopathologic study based on the new International Association for the Study of Lung Cancer/American Thoracic Society/European Respiratory Society international multidisciplinary lung adenocarcinoma classification. J Thorac Oncol. 2011;6:1496-1504. 
\item Nitadori JI, et al. Impact of micropapillary histologic subtype in selecting limited resection vs lobectomy for lung adenocarcinoma of 2cm or smaller. J Natl Cancer Inst. 2013;105:1212-1220.
\item Cha MJ, et al. Micropapillary and solid subtypes of invasive lung adenocarcinoma: clinical predictors of histopathology and outcome. J Thorac Cardiovasc Surg. 2014;147:921-928.
\item Woo T, et al. Prognostic value of the IASLC/ATS/ERS classification of lung adenocarcinoma in stage I disease of Japanese cases. Patho Int. 2012;62:785-791.
\item Tsao MS, et al. Subtype classification of lung adenocarcinoma predicts benefit from adjuvant chemotherapy in patients undergoing complete resection. J Clin Oncol. 2015;33:3439-3436.
\item Travis WD, Brambilla, E, Muller-Hemelink HK, Harris CC. World Health Organization Classification of Tumours. Pathology and Genetics of Tumors of the Lung, Pleura, Thymus, and Heart. Geneva: WHO Press; 2004.
\item Warth A, et al. Training increases concordance in classifying pulmonary adenocarcinomas according to the novel IASLC/ATS/ERS classification. Virchows Arch. 2012;461:185-193. 
\item Girard N, et al. Comprehensive histologic assessment helps to differentiate multiple lung primary non-small cell carcinomas from metastases. Am J Surg Pathol. 2009;33:1752-1764.
\item Miyoshi T, et al. Early-stage lung adenocarcinomas with a micropapillary pattern, a distinct pathologic marker for a significantly poor prognosis. Am J Surg Pathol. 2003;27:101-109.
\item Warth A, et al. Interobserver variability in the application of the novel IASLC/ATS/ERS classification for pulmonary adenocarcinomas. Eur Respir J. 2012;40:1221-1227.
\item Thunnissen E, et al. Reproducibility of histopathological subtypes and invasion in pulmonary adenocarcinoma. Mod Pathol. 2012;25:1574-1583.
\item Korbar B, et al. Deep learning for classification of colorectal polyps on whole-slide images. J Pathol Inform. 2017;8:30.
\item Tomita N, Cheung Y, Hassanpour S. Deep neural networks for automatic detection of osteoporotic vertebral fractures on CT scans. Comp Biol Med. 2018;98:8-15.
\item Litjens G, et al. A survey on deep learning in medical image analysis. Med Image Anal. 2017;42:60-88.
\item Coudray N, Moreira AL, Sakellaropoulos T, Fenyo D, Razavian N, Tsirigos A. Classification and mutation prediction from non-small cell lung cancer histopathology images with deep learning. Nat Med. 2018;24:1559-1567.
\item LeCun Y, Cortes C. MNIST handwritten digit database. 2010. Available from {\small \url{https://yann.lecun.com/exdb/mnist}}.
\item Netzer Y, Wang T, Coates A, Bissacco A, Wu B, Ng AY. Reading Digits in Natural Images with Unsupervised Feature Learning. NIPS. 2011.
\item Krasin I, Duerig T, Alldrin N, Andreas Veit. OpenImages: a public dataset for large-scale multi-label and multi-class image classification. 2017. Available from {\small \url{https://storage.googleapis.com/openimages/web/index.html}}.
\item Wang S, et al. Comprehensive analysis of lung cancer pathology images to discover tumor shape and boundary features that predict survival outcome. Sci Rep. 2018;10393.
\item Lou X, et al. Comprehensive computational pathological image analysis predicts lung cancer prognosis. J Thorac Oncol. 2017;12:501-509.
\item Yu KH, et al. Predicting non-small cell lung cancer prognosis by fully automated microscopic pathology image features. Nat Commun. 2016;7:12474.
\item Ren S, He K, Girshik R, Sun J. Faster R-CNN: towards real-time object detection with regional proposal networks. CVPR. 2015;91-99.
\item He K, Gkioxari G, Dollar P, Girshick R. Mask R-CNN. ICCV. 2017;2980-2988.
\item Shim HS, Lee DH, Park EJ, Kim SH. Histopathologic characteristics of lung adenocarcinomas with epidermal growth factor receptor mutations in the international association for the study of lung cancer/American thoracic society/European respiratory society lung adenocarcinoma classification. Arch Pathol Lab Med. 2011;135:1329-1334.
\item Levy M, et al. Histologic grade is predictive of incidence of epidermal growth factor receptor mutations in metastatic lung adenocarcinoma. Med Sci. 2017;5(4). 
\item Kumar N, et al. Identifying associations between somatic mutations and clinicopathologic findings in Lung Cancer Pathology Reports. Methods Inf Med. 2018;57:63-73.
\item Kadota K, Yeh YC, D'Angelo SP, Moreira AL, Kuk D, Sima CS. Associations between mutations and histologic patterns of mucin in lung adenocarcinoma: invasive mucinous pattern and extracellular mucin are associated with KRAS mutation. Am J Surg Pathol. 2014;38:1118-1127. 
\item He K, Zhang X, Ren S, Sun J. Deep residual learning for image recognition. CVPR. 2016;770-778.
\item Russakovsky O, et al. ImageNet large scale visual recognition challenge. IJCV. 2015;115:211-252.
\item Lin TY, et al. Microsoft COCO: common objects in context. ECCV. 2014;8693:740-755.
\item He K, Zhang X, Ren S, Sun J. Delving deep into rectifiers: surpassing human-level performance on ImageNet classification. ICCV. 2015;1026-1034.
\item McHugh, M. Interrater reliability: the kappa statistic. Biochem Med. 2012;22(3):276-82.
\end{enumerate}

\end{document}